\pdfoutput=1

\documentclass[11pt]{article}

\usepackage{acl}

\usepackage{times}
\usepackage{latexsym}
\usepackage{booktabs}
\usepackage{multirow}
\usepackage{array}
\usepackage{tikz}
\usepackage{float}
\usepackage{graphicx}
\usepackage{subcaption}
\usepackage{amsmath}
\usepackage[normalem]{ulem}
\usepackage{colortbl}
\usepackage{caption}
\usepackage{enumitem}
\usepackage{xcolor}

\setlist[enumerate]{itemsep=-1.5mm}

\usepackage[T1]{fontenc}

\usepackage[utf8]{inputenc}

\usepackage{microtype}

\newcommand{\taskname}{translation error correction}
\newcommand{\tasknameabr}{TEC}

\newcommand{\src}{$s$}
\newcommand{\pert}{$t$}
\newcommand{\tgt}{$t'$}
\newcommand{\datasetsize}{35,261}
\newcommand{\numdatasets}{three}
\newcommand{\datasetname}{\textsc{Aced}}
\newcommand{\asics}{\textsc{Asics}}
\newcommand{\digitalocean}{\textsc{DO}}
\newcommand{\emerson}{\textsc{Emerson}}
\newcommand{\modelname}{TEC}
\newcolumntype{P}{>{\fontfamily{lmss}\small}p{.7\textwidth}}
\newcolumntype{A}{>{\fontfamily{lmss}\small}p{.8\textwidth}}
\newcommand{\secref}[1]{Section~\ref{#1}}
\newcommand{\tabref}[1]{Table~\ref{#1}}
\newcommand{\figref}[1]{Figure~\ref{#1}}
\renewcommand{\approx}{\raise.17ex\hbox{$\scriptstyle\mathtt{\sim}$}}

\newcommand{\remove}[1]{\mybox{gitred}{\sout{#1}}}

\newcommand{\add}[1]{\mybox{gitgreen}{#1}}

\newcommand*{\V}[1]{\mathbf{#1}}
\newcommand*{\M}[1]{\mathbf{#1}}

\newif\ifcomments
\commentstrue 
\ifcomments
    \providecommand{\jessy}[1]{{\protect\color{blue}{[Jessy: #1]}}}
    \providecommand{\joern}[1]{{\protect\color{teal}{[Joern: #1]}}}
    \providecommand{\john}[1]{{\protect\color{orange}{[John: #1]}}}
    \providecommand{\geza}[1]{{\protect\color{purple}{[Geza: #1]}}}
\else
    \providecommand{\jessy}[1]{}
    \providecommand{\joern}[1]{}
    \providecommand{\john}[1]{}
    \providecommand{\geza}[1]{}
\fi

\newif\ifanonymize
\ifanonymize
    \providecommand{\lilt}{\textit{[anonymized]}}
\else
    \providecommand{\lilt}{Lilt}
\fi

\definecolor{gitred}{HTML}{FDB8C0}
\definecolor{gitgreen}{HTML}{ACF294}
\newcommand*{\mybox}[2]{\tikz[anchor=base,baseline=0pt,rounded corners=0pt, inner sep=0.2mm] \node[fill=#1] (X) {#2};}
\newcommand*{\graybox}[1]{{\color{gray}\scriptsize {#1}}}

\title{Automatic Correction of Human Translations}

\author{
Jessy Lin$^{\diamondsuit}$ \hspace{0.4cm} Geza Kovacs$^{\clubsuit}$ \hspace{0.4cm} Aditya Shastry$^{\clubsuit}$ \\
\textbf{Joern Wuebker$^{\clubsuit}$ \hspace{0.4cm} John DeNero$^{\diamondsuit\clubsuit}$} \\
$^{\diamondsuit}$ University of California, Berkeley \hspace{0.4cm} $^{\clubsuit}$ Lilt\\
\texttt{jessy\_lin@berkeley.edu, joern@lilt.com, john@lilt.com}}

\begin{document}
\maketitle
\begin{abstract}
We introduce translation error correction (TEC), the task of automatically correcting human-generated translations.
Imperfections in machine translations (MT) have long motivated systems for improving translations post-hoc with automatic post-editing.
In contrast, little attention has been devoted to the problem of automatically correcting human translations, despite the intuition that humans make distinct errors that machines would be well-suited to assist with, from typos to inconsistencies in translation conventions.
To investigate this, we build and release the \datasetname{} corpus with \numdatasets{} TEC datasets\footnote{Dataset available at: \url{https://github.com/lilt/tec}}. We show that human errors in TEC exhibit a more diverse range of errors and far fewer translation fluency errors than the MT errors in automatic post-editing datasets, suggesting the need for dedicated TEC models that are specialized to correct human errors. We show that pre-training instead on synthetic errors based on human errors improves TEC F-score by as much as 5.1 points.
We conducted a human-in-the-loop user study with nine professional translation editors and found that the assistance of our TEC system led them to produce significantly higher quality revised translations.
\end{abstract}

\section{Introduction}

\begin{table*}[ht]
\centering
\begin{tabular}{>{\bfseries}lP}\toprule
\textbf{Error Type} & \multicolumn{1}{l}{\textbf{Example Text}} \\\hline
\multirow{3}{*}{Monolingual: typos} & \src{}: Do your feet roll inwards when running? \\
& \pert{}: K\mybox{gitred}{I}ppen deine Füße beim Laufen nach innen? \\
& \tgt{}: K\mybox{gitgreen}{i}ppen deine Füße beim Laufen nach innen? \\
\hline
\multirow{3}{*}{Monolingual: grammar} & \src{}: Own tough winter runs in the \ldots \\
& \pert{}: Bei harten Winterläufe sorgt der \ldots \\
& \tgt{}: Bei harten Winterläufe\mybox{gitgreen}{n} sorgt der \ldots \\
\hline
\multirow{3}{*}{Monolingual: fluency} & \src{}: The traffic emerges from the VPN server and \ldots \\
& \pert{}: Der \mybox{gitred}{Verkehr} wird vom VPN-Server ausgegeben und \ldots \\
& \tgt{}: Der \mybox{gitgreen}{Datenverkehr} wird vom VPN-Server ausgegeben und \ldots \\
\hline
\multirow{3}{*}{Bilingual} & \src{}: Quad Core XEON E3-1501M, 2.9GHz \\
& \pert{}: Quad Core XEON 2,9 GHz \\
& \tgt{}: Quad Core XEON \mybox{gitgreen}{E3-1501M}, 2,9 GHz \\
\hline
\multirow{3}{*}{Preferential} & \src{}: VersaMax I / O auxiliary spring clamp style \\
& \pert{}: VersaMax Zusatz-E / A \mybox{gitred}{Federklemmenart} \\
& \tgt{}: VersaMax Zusatz-E / A \mybox{gitgreen}{Federklemmenbauform} \\
\bottomrule
\end{tabular}
\caption{Error taxonomy for the \datasetname{} corpus, with examples from the dataset.}
\label{tab:error-taxonomy}
\vspace{-4mm}
\end{table*}

Despite recent progress in machine translation (MT), a tremendous amount of translated content in the world is still written by humans \citep{depalma-2021-csa}. Humans are often assumed to produce trusted, high-quality translations. In reality, they do make errors, including spelling, grammar, and translation errors \citep{hansen2009errors}. This paper introduces the task of \textbf{\taskname{}} (\tasknameabr{}). Given a source sentence \src{} and a human-generated translation \pert{}, the goal of \tasknameabr{} is to produce an improved translation \tgt{} by correcting all errors in \pert{}.

``Translation correction'' has long been studied in the MT community through the task of automatic post-editing (APE), which aims to correct errors in machine-generated translations \citep{simard-etal-2007-rule}. \tasknameabr{} is structurally identical to APE. However, it requires modeling a different data distribution: errors made by humans, which differ from those made by MT systems \citep{freitag2021experts}. We characterize the error distribution in TEC by building, analyzing, and releasing the \datasetname{} corpus, a collection of \numdatasets{} TEC datasets from varying domains, with a total of \datasetsize{} English–German translations produced and corrected by professional translators in the natural course of their work. While APE is dominated by the fluency errors that are characteristic of MT systems (74\% of sentences), our TEC corpus exhibits a broader distribution of errors that human translators are prone to make.

Using this error analysis, we propose an approach for TEC that pre-trains on synthetic corruptions more similar to errors made by humans, outperforming models that were developed for the related tasks of MT, grammatical error correction, and APE on all \datasetname{} datasets.

The task of TEC is often currently performed by humans, e.g. translators hired to review and edit translations (``reviewers''). Can a TEC system help reviewers edit faster, or produce higher quality final translations than they would have without assistance? We ran a human-in-the-loop user study with nine professional translators using our best-performing TEC model. We found that the reviews produced when assisted with a TEC system were rated as higher quality than those produced without, and produced with less manual effort. Qualitatively, users commented that trust and consistency of the suggestions were critical. They speculated future automated assistance could be helpful for onboarding to new content, spotting technical errors, and improving their own awareness of errors to catch.

Looking forward, a natural question arises of whether the research community should focus on learning to revise \emph{model} outputs (APE) or \emph{human} outputs (TEC). With recent improvements in MT, it has been increasingly difficult for APE models to improve model output that is already high quality \citep{chollampatt2020automatic}. On the other hand, we should expect that humans will continue to make errors. \tasknameabr{} models will continue to provide benefit as a way of assisting humans, whether for professional translators or everyday language learners. \tasknameabr{} is synergistic with continuing advancements in MT: improved MT will lead to improved error correction for human-generated translations. While APE pits models against models, \tasknameabr{} is an opportunity to combine the best of humans and models \emph{because} humans and models make different errors.

In sum, this paper revisits the notion of ``translation correction'' conceived narrowly as the MT-centered task of APE, with an empirical investigation of translation error correction (TEC), the task of learning to correct human translations. Our contributions are:
\vspace{-1mm}
\begin{enumerate}
    \item We release \datasetname{}, the first corpus for TEC containing \numdatasets{} datasets of human translations and revisions generated naturally from a commercial translation workflow.
    \item We analyze the kinds of errors humans make in \datasetname{}, finding that while APE is dominated by correcting translation \emph{fluency}, TEC focuses on correcting a broader range of errors that appear in translation.
    \item We propose a pre-training approach for TEC that outperforms approaches developed for similar tasks such as APE. Together, our results suggest the need for distinct approaches to correct human translation errors.
    \item We perform a human-in-the-loop user study, finding that professional translators produce \textit{higher quality} translations when assisted by a TEC model.
\end{enumerate}
\section{The $\spadesuit$ \datasetname{} Corpus for TEC}\label{sec:task-description}

\begin{table*}[t]
\centering
\small
\begin{tabular}{@{}lccccccc@{}}\toprule
\multirow{2}{*}[-.25em]{\bf Dataset} & \multirow{2}{*}[-.25em]{Domain} & \multicolumn{3}{c}{\# sentences} & \multirow{2}{*}[-.25em]{\% edited}  & \multirow{2}{*}[.1em]{\# edits} & \multirow{2}{*}[.1em]{Edit distance} \\
\cmidrule{3-5}
& & train & dev & test & & (mean) & (mean) \\
\midrule
\asics{}         & Marketing & \phantom{0}1395 & \phantom{0}525 & \phantom{0}616          & 29 \%         & 1.6           & 7.5   \\
\emerson{}      & Technical & \phantom{0}4287 & 1255 & 1662        & 20 \%         & 1.5           & 5.8 \\
\textsc{DigitalOcean (DO)}      & Technical & 11773 & 6104 & 7644       & \phantom{0}8 \%   & 1.5       & 7.1  \\ 
\bottomrule
\end{tabular}
\caption{Corpus statistics for each dataset in the \datasetname{} corpus, including edit statistics on \% of sentences that edited, and for edited sentences, the average number of edits and average edit distance.}
\label{tab:basic-corpus-stats}
\end{table*}

Given a source language sentence \src{} and a human-generated translation \pert{}, the goal of TEC is to produce a corrected target language sentence \tgt{}.

We introduce the \datasetname{} corpus, a set of \numdatasets{} TEC datasets: \asics{}, \emerson{}, and \textsc{DigitalOcean} (\digitalocean{}), each consisting of English–German sentence triples (\src, \pert, \tgt) from varying domains.

\datasetname{} is a \emph{real-world benchmark}, containing naturalistic data from a task humans perform, rather than manually annotated data. All translations were created by professional translators working with \lilt{}, a localization services provider. All translators have at least 5 years of professional translation experience and experience working with the customer and domain. Each document was translated from scratch (i.e. not post-edited) by a human translator using an interactive neural MT system. Each translated document was then reviewed by a reviewer, who \lilt{} selects as one of the more senior translators. As a result, the examples in our corpus exhibit real errors that translators make, and the corrected translations are publication quality. 

Secondly, \datasetname{} is \emph{diverse}, with the \numdatasets{} datasets from varying domains exhibiting different error distributions and difficulty for initial work on the TEC task. Information for each dataset is shown in \tabref{tab:basic-corpus-stats}. The \asics{} dataset consists of marketing content with product names and descriptions for an activewear company. \emerson{} consists of industrial product names for a manufacturing company. \digitalocean{} consists of software engineering tutorials. The various content types pose different challenges for translators and thus for TEC systems, which we discuss in \secref{sec:tec-ape-errors} and \secref{sec:edit-overlap}.

Duplicate sentences with the same source \src{} were removed. A portion  of sentences were rewritten by the reviewer rather than being edited (a relative edit distance of more than 25\% and a minimum of two edited words). We replace \pert{} in these sentences with \tgt{} in the corpus, so that training and evaluation focuses on local edits rather than re-translations. The train, dev, and test splits were constructed by splitting along document boundaries.

\subsection{How do TEC and APE errors differ?}
\label{sec:tec-ape-errors}
To understand how the human errors in the TEC task differ from model errors in APE, we compare the types of errors in \datasetname{} with 100 randomly sampled errors in the WMT 2021 APE shared task dev set~\footnote{\texttt{\scriptsize{https://www.statmt.org/wmt21/ape-task.html}}}, which we then annotate with error types. We define an error taxonomy that classifies each edit as one of three types: (1) \textbf{Monolingual edits} are identifiable from only the target-side text. We divide these further into subcategories that highlight different capabilities needed to correct edits: \emph{typos} (including spelling, punctuation, spacing, orthographic issues), \emph{grammar}, and \emph{fluency} (awkward phrasing, word choice, or non-native-sounding disfluencies); (2) \textbf{Bilingual edits} concern mismatches between the source and target text, e.g. over- or under-translation, mis-translations; (3) \textbf{Preferential edits} correct text that is inconsistent with the preferences of the customer, as described in extralinguistic project requirements (e.g. terminology or stylistic preferences). Examples of each error type are shown in \tabref{tab:error-taxonomy}. Our error taxonomy closely mirrors those of previous analyses of human translation errors \citep{specia-shah-2014-predicting, yuan-sharoff-2020-sentence, Gupta2021DetectingOE}, and we confirm their findings that human translation errors differ from MT errors. However, while previous work focuses on error \emph{detection} and quality estimation, TEC is concerned with error \emph{correction}. Our error types are intended to isolate the capabilities that models need to learn to correct edits (e.g., target-side language models can learn to correct monolingual errors, but cannot do well on bilingual edits).

We annotate and release error labels for all test sentences in \asics{} to enable its use as a diagnostic set for per-type evaluation of models. On the larger \emerson{} and \digitalocean{} datasets, we randomly sample 50 errors to annotate for this analysis. Error types were annotated by a professional German translator. Each segment can have multiple error types. In \tabref{tab:error-breakdown}, we report the percentage of sentences with at least  one error of each type. 74\% of sentences in APE exhibit a fluency error, in contrast to up to 22\% of sentences in \datasetname{}, while other types like monolingual grammar, bilingual, and preferential errors are notably underrepresented in APE. We also note that all sentences in the APE shared task are edited, while a key feature of TEC is identifying when a sentence does \emph{not} need to be edited. The error distributions suggest that different modeling techniques may shine: while APE challenges models to correct disfluent translations characteristic of MT systems, our task is designed to focus on identifying and correcting the typos, mismatches, and grammatical errors more commonly exhibited by humans. Guided by this observation, we describe an approach in \secref{sec:baselines} that pre-trains on synthetic edits that are more representative of this error distribution.

\subsection{How difficult is it to learn to edit?}
\label{sec:edit-overlap}
\begin{table}[t]
\centering
\small
\begin{tabular}{lccc|c}\toprule
& \multicolumn{3}{c}{TEC} & APE \\
& \asics{} & \emerson{} & \digitalocean{} & WMT '21 \\
\midrule
Monolingual & & & & \\
\hspace{1.5mm} typos & 13 & 16 & 22 & 16 \\
\hspace{1.5mm} grammar  & 41 & \phantom{0}4 & \phantom{0}2 & \phantom{0}6 \\
\hspace{1.5mm} fluency  & 22 & \phantom{0}0 & 20 & 74 \\
Bilingual             & 22 & 70 & 32 & \phantom{0}5 \\
Preferential          & \phantom{0}7  & 24 & 40 & \phantom{0}6 \\
\bottomrule
\end{tabular}
\caption{Percentages of erroneous sentences that contain at least one error of each type for \asics{}, \emerson{}, \digitalocean{}, and the dev set of the WMT 2021 APE shared task. As a task, APE exhibits many more fluency errors than TEC.}
\label{tab:error-breakdown}
\vspace{-3mm}
\end{table}

To quantify how difficult it may be to learn the correct edits in  \datasetname{}, we report statistics on \emph{edit overlap}: what proportion of edits that we expect models to perform (e.g. adding a comma) appear exactly in the training set? We use the \textit{errant} toolkit to identify discrete edits  \cite{bryant-etal-2017-automatic}. Each edit is represented as a tuple \textit{(original span, replacement span)}, e.g. \textit{(``auf'', ``an'')} to replace ``auf'' with ``an.'' Edit statistics are reported in \tabref{tab:edit-overlap}: in \asics{} and \digitalocean{}, \approx{}20\% of the total number of edits in dev and test appear in the training set, while \emerson{} has \approx{}60\% of dev and test edits appear in the training set.

While the edit overlap rate provides a relative sense of scale for precision and recall numbers, it does not provide an upper bound on recall. It is possible to learn edits that do not exactly appear in the training set. For example, capitalizing product names (``winterized'' $\rightarrow$ ``WINTERIZED'') is a learnable pattern that would appear as many distinct edits. Additionally, some errors can be corrected without fine-tuning because they are generic typo, grammatical, fluency, or bilingual errors. Conversely, it is also possible that it is wrong to make an edit that appears in the training set, depending on the surrounding sentential context.
\begin{table}[t]
\centering
\small
\begin{tabular}{llccc}\toprule
 & & \asics{} & \emerson{} & \digitalocean{} \\
\midrule
\multirow{2}{*}{Train} & Total Edits & 606 & 1436 & 1212 \\
                        & Unique Edits & 418 & \phantom{0}486 & \phantom{0}940 \\
\midrule
\multirow{2}{*}{Dev}  & Total Edits & 246 & \phantom{0}381 & 1004\\
                    & \% in train & \phantom{0}14 & \phantom{0}\phantom{0}63 & \phantom{00}21 \\
\midrule
\multirow{2}{*}{Test}  & Total Edits & 287 & \phantom{0}364 & \phantom{0}766 \\
                    & \% in train & \phantom{0}23 & \phantom{00}60 & \phantom{00}21 \\
\bottomrule
\end{tabular}
\caption{Edit and edit overlap statistics for each \datasetname{} dataset: total number of edits, unique edits in each train split, and percentage of total edits in each dev and test split that appear in the train split.}
\label{tab:edit-overlap}
\end{table}
\section{Approaches to TEC}
\label{sec:baselines}
\begin{figure*}[ht]
  \centering
  \includegraphics[width=.85\textwidth]{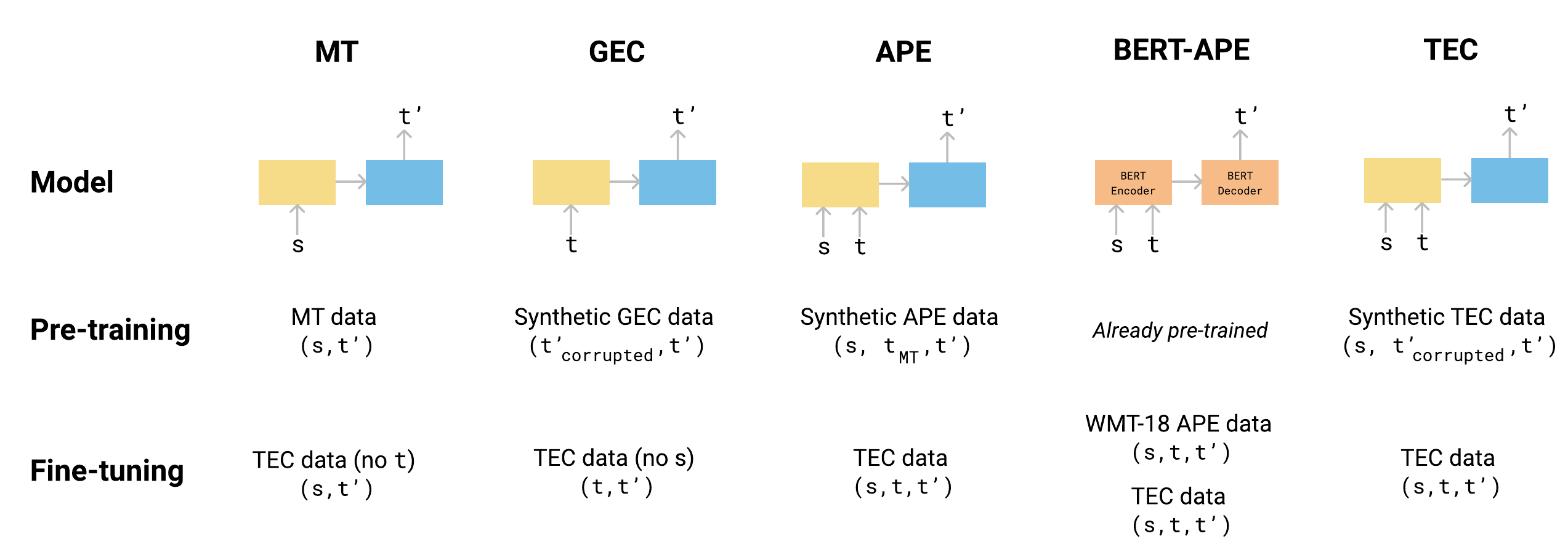}
  \caption{Overview of model architectures, pre-training data, and fine-tuning data for each approach to TEC. Transformer encoders and decoders are depicted as yellow and blue rectangles, respectively. The GEC and \tasknameabr{} models are pre-trained with synthetic corruptions of \tgt{} ($t'_{\text{corrupted}}$), as detailed in the description of the TEC and GEC models. APE uses MT-generated translations of \src{} ($t_{\text{MT}}$) as synthetic data. BERT-APE is a state-of-the-art pre-trained APE model made available by \citet{correira2019apebert}.}
  \label{fig:models}
\end{figure*}

We propose a \tasknameabr{} model and compare it to several models designed for related tasks to determine whether they are also effective for \tasknameabr{}. An overview of the differences between the models is shown in Figure~\ref{fig:models}. All models use the Transformer neural architecture \citep{vaswani17transformer} that generates the target sequence \tgt{} from left to right. All are pre-trained on 36M sentences from the WMT18 translation task\footnote{\url{http://www.statmt.org/wmt18}} and fine-tuned on \datasetname{}, unless indicated otherwise. All pre-training and fine-tuning data is pre-processed by normalizing punctuation with the Moses toolkit \cite{koehn-etal-2007-moses}.

All models have 6 encoder and decoder layers, model dimension of 256, feed-forward dimension of 512, and 8 attention heads. We use a joint English–German vocabulary with 33k byte pair encoding subwords \citep{sennrich-2016-bpe}. During pre-training, we set the dropout to 0.1 and use the Adam optimizer \citep{kingma-2015-adam} with a learning rate of $0.0002$. During fine-tuning, we decrease the learning rate to $0.0001$ and reset the Adam momentum parameters. We select the best fine-tuning checkpoint with edit-level F$_{0.5}$ score on our dev set. We use greedy inference.

\subsection{Dual-Source Encoder-Decoder Model}
\label{sec:dual-source}
We first describe the dual-source encoder-decoder we use for the APE and TEC models. Formally, the original Transformer architecture \citep{vaswani17transformer} takes a sequence of $J$ source tokens $\V{s}_{1 \dots J}$ and predicts a sequence of $I'$ target tokens $\V{t'}_{1 \dots I'}$. We adapt the architecture to additionally encode the original translation \pert{}, a sequence of $I$ tokens, $\V{t}_{1 \dots I}$. We independently project $\V{t}$ into the embedding space, add an offset vector $\V{o}$, and then concatenate the embedding with the embedding of the source $\V{s}$ to form the encoder input. To allow the dual-source model to copy tokens from the original translation \pert{}, we implement the copy-mechanism proposed by \citet{zhao2019copygec}, which augments the model with an additional encoder-decoder attention layer. An expanded description of the model can be found in Appendix \secref{sec:appendix-model-description}.

\subsection{Synthetic Data Generation}
\label{sec:synth-data}
For the TEC and GEC model, we generate synthetic triples (\src{}, \pert{}, \tgt{}) for pre-training. We generate a synthetic $t$ by corrupting the German side of the translation data into $t'_{\text{corrupted}}$. For each sentence, we sample the probability of corruption $p_c \sim \mathcal{N}(\mu=0.01, \sigma=0.04)$ clipped at $0$. On each character and word in that sentence, with probability $p_c$, we randomly select one of the following perturbations to apply at that position: insertion, deletion, transposition, repetition.

\subsection{TEC Models}
\begin{table*}[t]
\centering
\small
\begin{tabular}{lccccccccc}\toprule
& \multicolumn{3}{c}{\asics{}} & \multicolumn{3}{c}{\emerson{}} & \multicolumn{3}{c}{\digitalocean{}} \\ 
\cmidrule(lr){2-4}\cmidrule(lr){5-7}\cmidrule(lr){8-10}
\bf Model & Prec. & Rec. & F$_{0.5}$ & Prec. & Rec. & F$_{0.5}$  & Prec. & Rec. & F$_{0.5}$ \\\midrule
MT      & 3.3 & \bf 31.4 & 4.0   & 16.2 & \bf 78.6 & 19.2   & 1.2	& \bf 41.9	& 1.5      \\
GEC     & 52.8 & 6.6 & 22.0    & 78.0 & 53.3 & 71.4     &  20.5  & 2.0 &	7.1      \\
APE     & 51.2 & 7.3 & 23.3    & 78.1 & 54.4 & 71.8   & 14.4 &	1.7	& 5.8  \\
BERT-APE & 6.8 & 10.8 & 7.3     & 32.0 & 57.8 & 35.1  & 2.3	 & 3.8	& 2.5 \\
\midrule
\modelname{} & \bf 57.4 & 9.4 & \bf 28.4  & \bf 82.1 & 57.2 & \bf 75.5 & \bf 21.7 &	2.0	& \bf 7.2\\
\bottomrule
\end{tabular}
\caption{Main results on \datasetname{}. Our fine-tuned \modelname{} model outperforms on F$_{0.5}$. The fine-tuned MT model scores highest on recall because it makes many edits, but at the cost of unacceptably low precision.}
\label{tab:results}
\end{table*}

The five approaches we compare are:

\paragraph{\modelname{} (this work)} We implement the dual-source encoder-decoder model that encodes two inputs (\src{}, \pert{}) and outputs \tgt{}, as described in \secref{sec:dual-source}, and then pre-train on synthetic data from the procedure in \secref{sec:synth-data}. We then fine-tune on \datasetname{}.

\paragraph{MT} We train an English-German neural machine translation model (with the standard architecture described previously) and fine-tune it on (\src{}, \tgt{}) \datasetname{} pairs, ignoring the original translation \pert{}.

\paragraph{GEC} We evaluate a encoder-decoder (monolingual) GEC model that takes an incorrect German sentence \pert{} as input and outputs a corrected \tgt{}. We use the same copy mechanism to attend to \pert{} as our \tasknameabr{} model. To pre-train, we perturb \tgt{} using the procedure described in \secref{sec:synth-data}, throwing away the source side to obtain $(t,t')=(t'_{\text{corrupted}},t')$ pairs. We then fine-tune on the \datasetname{} corpus, ignoring \src{}.

\paragraph{APE} We implement a dual-source encoder-decoder model that is identical to our \tasknameabr{} model. Following common practice in APE \citep{junczys-dowmunt-2016-logape, negri-etal-2018-escape}, we pre-train on synthetic ``post-editing'' triples (\src{}, \pert{}, \tgt{}) where $t=t_{\text{MT}}$ is generated by translating \src{} with an MT system. We split the training dataset into two parts, train an MT model on each half, and use each model to translate the other half of the dataset not seen during training. We then fine-tune on \datasetname{}.

\paragraph{BERT-APE} We also evaluate whether a state-of-the-art APE model can be directly applied to our task. We evaluate the BERT-based encoder-decoder of \citet{correira2019apebert}, on which the WMT 2019 shared task winner was based \citep{lopes-2019-unbabelbert}. Following \citet{correira2019apebert}, we fine-tuned on 23K English–German SMT triplets from the WMT18 shared task\footnote{\url{http://www.statmt.org/wmt18/ape-task.html}, with 12K train triplets from 2016 and 11k from 2017}. We reproduce their results on the APE shared task test sets, and continue fine-tuning this model on \datasetname{}. Following their paper, the inputs are pre-processed by tokenizing and joining the two inputs with a separator to form (\src{} \texttt{[SEP]} \pert{}, \tgt{}) pairs.

\section{Results \& Discussion}
\label{sec:analysis}

The primary metric for \tasknameabr{} is MaxMatch scores (M$^2$) \citep{dahlmeier2012m2} computed with the \textit{errant} toolkit \citep{bryant-etal-2017-automatic}. M$^2$ is a standard metric for GEC that aligns \pert{} and \tgt{} to extract discrete ``edits.'' We choose to follow the GEC evaluation practice of up-weighting precision by comparing F$_{0.5}$, since the original translation is mostly correct: it is better to suggest few correct edits than potentially introduce new errors.

Table~\ref{tab:results} shows that \modelname{} achieves the best overall F$_{0.5}$ score on all datasets, from $+0.1$ (on \digitalocean{}) up to $+5.1$ (on \asics{}) above the next-best model. Fine-tuning on actual human corrections provides substantial gains; results without fine-tuning can be found in Appendix~\secref{sec:appendix-base-results}.

Both the MT model (which ignores \pert{}) and the GEC model (which ignores \src{}) underperform \modelname{}. The MT model's high edit recall can be attributed to the fact that it proposes many edits, greatly trading off precision. Without conditioning on \pert{}, direct MT translations of the source diverge from the reference. The GEC model obtains high precision but underperforms on recall. Conditioning on \src{} not only makes it possible to propose bilingual edits, but also provides additional information to correct monolingual edits, as we show in \secref{sec:per-cat-analysis}.

\paragraph{Can APE models be directly adapted for \tasknameabr{}?}
Since TEC is structurally identical to APE, a natural question is whether models that are trained on the APE objective can be directly adapted for \tasknameabr{}. The APE and \modelname{} models differ only in pre-training, but the performance difference between them is substantial, indicating that the more GEC-like data synthesis procedure is a better fit for TEC. Even more, the BERT-APE model, which is first fine-tuned to achieve state-of-the-art on APE before fine-tuning on \datasetname{}, achieves a particularly low F$_{0.5}$ score because it makes too many edits (low precision). Although future work may find insights in APE, our results emphasize that models that excel at correcting machine errors cannot be assumed to work well on human translations.

\subsection{Fluency \& Per-category Error Analysis}
\label{sec:per-cat-analysis}
\begin{table*}[t]
\centering
\small
\begin{tabular}{lcccccccc}\toprule
& & \multicolumn{7}{c}{\bf Sentence-level Accuracy (\%)} \\
\cmidrule{3-9}
 & & \multirow{2}{*}{\it Overall} & \multirow{2}{*}{Unedited} & Mono. & Mono. & Mono. & \multirow{2}{*}{Bilingual} & \multirow{2}{*}{Preferential} \\
 & & & & Typos & Grammar & Fluency &  &  \\
\bf Model & \bf GLEU & \graybox{/616} & \graybox{/435} & \graybox{/16} & \graybox{/78} & \graybox{/41} & \graybox{/41} & \graybox{/14}\\\midrule
No-edit & 87.85 & 70.62 & (100) & - & - & - & - & - \\ 
MT & 44.79 & 31.66 & 40.00 & \phantom{0}0.00 & 11.54 & \bf 2.44 & \bf 24.39 & \bf 14.29 \\ 
GEC & 88.46 & 71.10 & 97.01 & \bf 31.25 & 14.10 & 0.00 & \phantom{0}0.00 & \phantom{0}0.00 \\ 
APE & 88.39 & 71.43 & \bf 97.47 & 12.50 & 16.67 & \bf 2.44 & \phantom{0}0.00 & \phantom{0}0.00 \\ 
BERT-APE & 82.35 & 39.77 & 52.18 & 6.25 & \bf 19.23 & 0.00 & \phantom{0}4.88 & \phantom{0}0.00 \\
\midrule
\modelname{} (this work) & \bf 88.81 & \bf 71.92 & 97.01 & 25.00 & 17.95 & 0.00 & \phantom{0}7.32 & \phantom{0}0.00 \\ 
\bottomrule
\end{tabular}
\caption{Additional analysis of n-gram overlap (via GLEU) and exact-match sentence accuracy over all test sentences (Overall) and per error category for the \asics{} dataset. Number of sentences with an error of each type are indicated in gray, with some sentences containing errors from multiple categories. For each error category, we report the percentage of those sentences with at least one error of that type that a system predicted exactly.}
\label{tab:results-aux}
\end{table*}

We perform a more in-depth comparison using alternative metrics on \asics{}, which includes annotated error labels as a diagnostic tool. First, to understand how much models are editing, we look at \emph{n-gram overlap} with the GLEU metric \citep{napoles2015gleu}, a variant of BLEU used in GEC evaluation to measure the fluency of holistic rewrites \citep{sakaguchi2016gleufluency}. Next, we compare \emph{sentence-level accuracy}, which measures exact match with \tgt{}. We compute overall sentence-level accuracy, which includes unedited sentences (which some models may incorrectly edit). We also report accuracy per error type over (edited) sentences annotated with that error type. These metrics need to be interpreted carefully: a no-edit baseline achieves a GLEU score of $87.85$ (since original translations are mostly close in edit distance to the final) and sentence-level accuracy of $70.62\%$ (the \% of unedited sentences), outperforming models like MT and BERT-APE that make too many incorrect edits.

Our \modelname{} model achieves the best score overall on both alternative metrics over all sentences, but various models outperform on specific error types. The full results are shown in \tabref{tab:results-aux}. Examples of system outputs for different error types can be found in Appendix \secref{sec:appendix-system-outputs}.

Notably, APE models lose the most accuracy relative to our model on monolingual typo edits. This may be because neural MT decoders much less frequently introduce target-side errors that would be similar to typos (compared to the frequency of fluency errors). Still, the observation that different models do well at different errors suggests that future work can improve on TEC by leveraging the strengths of different models, e.g. using MT models to propose alternative translations.

\section{User Study: Assisting Professional Translators with TEC}
\label{sec:user-study}
Our automatic evaluation shows how our TEC model can outperform other baseline systems, but we are ultimately interested in whether any TEC system is indeed useful in practice. Presently, TEC is done manually by humans. To investigate whether TEC systems can already be useful to humans---improving the quality, speed, or ease of human review---we performed a human-in-the-loop user study with our TEC model.

\subsection{Methodology}



We recruited 9 professional translators to serve as reviewers. None of them had prior experience with \asics{} content. They were allowed to read and reference the sentences in the \asics{} training set to familiarize themselves with the content and preferred terminology. Then, they were each assigned to review 74 sentences from the test set of \asics{}. Of the 74 sentences, our TEC system predicted a suggested edit for 57 sentences, and for the remaining 17 sentences our TEC system did not predict any edits.

We opt for a within-subjects design to control for speed and experience differences between reviewers. For each reviewer, the 74 sentences were randomized such that half were in the ``assisted condition'' showing the TEC suggestion if available for the sentence, and the other half were in the ``unassisted condition'' where no TEC suggestion was shown.
\begin{figure}[t]
  \includegraphics[width=\linewidth]{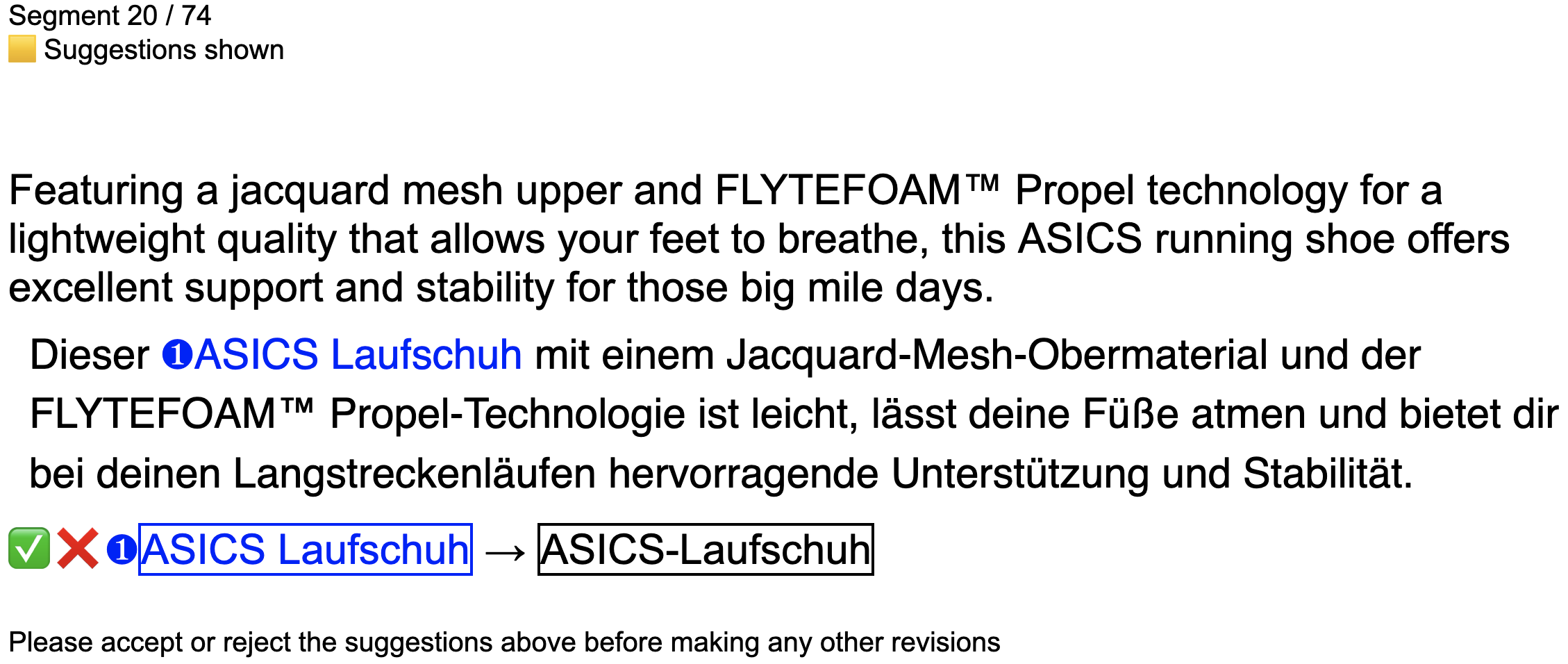}
  \caption{The interface used to show suggestions to reviewers in our user study.}
  \label{fig:review-interface}
  \vspace{-2mm}
\end{figure}
The reviewing interface is shown in \figref{fig:review-interface}. If the sentence has suggestions available, the reviewer is asked to first accept or reject each of the suggestions. Then, they are asked to make edits to the text until they are satisfied with the translation. They then click a button to confirm their translation and move to the next sentence.

During the review process, we track:
\vspace{-1mm}
\begin{enumerate}
    \item Whether the TEC suggestion, if shown, was accepted or declined
    \item Total time spent reviewing each sentence
    \item Number of edit operations (insertions and deletions) the user made
    \item Levenshtein edit distance from the original text to the final text
\end{enumerate}
\vspace{-1mm}
\begin{figure*}[!t]
\noindent
\begin{minipage}[t]{0.49\textwidth}
\vspace{0pt}
\centering
\small
\begin{tabular}{@{}lccc@{}}
\toprule
        & Review Time & Inserts & Levenshtein \\
\bf Suggestions & (ms/char) & + Deletes & Distance \\
\midrule
Hidden & 361 & 0.0625 & 0.0347 \\
Shown \textit{(Overall)} & 367 & 0 & 0.0185 \\
\hspace{4mm}\textit{Accepted} & 328 & 0 & 0.0176 \\
\hspace{4mm}\textit{Declined} & 841 & 0.0625 & 0.0426 \\ \bottomrule
\end{tabular}
\captionof{table}{Length-normalized medians from our user study for review times, number of characters the user inserted and deleted, and final Levenshtein edit distances from the original to the final translations. Data is split based on whether the suggestion was hidden or shown, and ``Suggestion Shown'' is further broken down according to whether the user accepted or declined the suggestion.}
\label{tab:user-study-results}
\end{minipage}%
\hfill%
\begin{minipage}[t]{0.49\textwidth}
  \vspace{0pt}
  \centering
  \includegraphics[width=\textwidth]{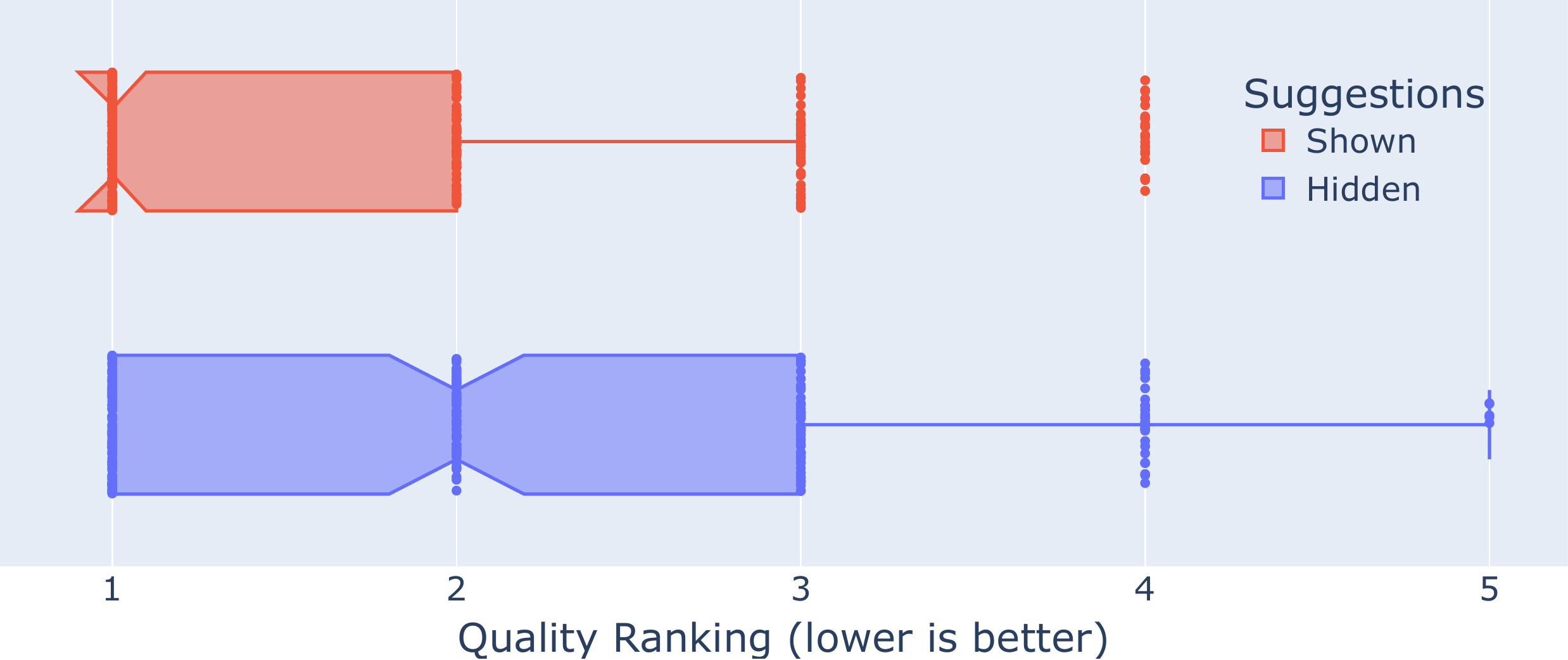}
  \captionof{figure}{Box plot showing quality rankings of segments reviewed with suggestions hidden vs shown. Lower is better. Notches indicate median quality rankings. Bars indicate the upper fence ($3^{rd}$ quartile + IQR*1.5).}
  \label{fig:quality-ranking}
\end{minipage}%
\end{figure*}

Finally, to evaluate whether TEC has an effect on quality, we asked a 10th translator to  compare the quality of the reviewed sentences by ranking the 9 reviewed translations, with ties allowed. This translator was the translator who had reviewed the reference translations in the corpus, as the explicit goal is to ensure consistency with conventions in the training documents.

\subsection{Results}

In the user study, \textbf{79\%} of TEC suggestions were accepted. For the purpose of analyzing effects of the TEC suggestions on time spent and translation quality, we will focus on only the 255 sentences (across 9 reviewers) where a TEC suggestion exists. Results are shown in \tabref{tab:user-study-results}. For all statistical significance tests, we use the Mann-Whitney (MW) U-test for testing statistical significance as all quantities are neither normally distributed nor log-normal.

\subsubsection{Effects of Suggestions on Time Spent During the Review Process}
\label{sec:user-study-time-spent}

We first analyze how TEC suggestions influence the time spent reviewing a sentence. We compare the time durations normalized by the length of the sentence that needed to be reviewed (the \textit{length-normalized review time}), as longer sentences require more time to be read and reviewed.  

There is no significant difference in length-normalized review time when the suggestion is hidden vs. shown (MW $U=31654, p=0.460$). When suggestions are shown, the length-normalized review time is significantly less on sentences where reviewers accepted the suggestion, compared to sentences where they declined (MW $U=3555, p<0.0005$).

A potential explanation for these results is that when reviewers are shown incorrect TEC suggestions, they are distracted and slowed down, providing some evidence that precision should indeed be emphasized in automatic evaluations of TEC.

\subsubsection{Effects of Suggestions on Edits Made During the Review Process}

We also analyze the effects of suggestions on the editing effort, as measured by the number of characters the reviewer had to insert and delete, as well as how different the final reviewed sentences were from the original.

When suggestions are shown vs. hidden, there is a significant reduction in the number of insertions+deletions (MW $U=41348.5, p<0.0001$). There is also a significant reduction when a shown suggestion is accepted vs. declined (MW $U=4007.5, p<0.005$). There is no significant difference in the Levenshtein distance from the original translation to the final translation, between when a suggestion is shown vs. hidden (MW $U=33750.0, p=0.611$), or between when a shown suggestion is accepted vs. declined (MW $U=5485.0, p=0.783$).

Thus, the TEC system suggestions help to significantly reduce the amount of manual typing that the user must perform.


\subsubsection{Effects of Suggestions on Reviewed Sentence Quality}
\vspace{-.5mm}
To assess the effects of TEC assistance on quality, we used the quality rankings produced by the independent reviewer. Quality rankings were not normally distributed, so we use the Mann-Whitney U-test for testing statistical significance. A box plot of the quality rankings is shown in \figref{fig:quality-ranking}. The median quality ranking when the suggestion is shown is 1, vs. 2 when the suggestion is hidden. The quality ranking is significantly lower (meaning quality is higher) when the suggestion is shown, vs. hidden (MW $U=28738.0, p<0.01$). 

This suggests that showing TEC suggestions may be helping reviewers correct errors they may not have otherwise noticed, or help nudge them towards desired corrections.

\subsection{Qualitative Findings}

We also conducted a post-study survey for reviewers to report qualitative feedback. To understand common themes in the responses, we present all themes that at least two reviewers mention in their commentary.
\vspace{-2mm}
\subsubsection{The Role of Reliability and Trust}
Five reviewers commented that \emph{reliability is critical}: it was difficult to trust the system when they noticed some suggestions were incorrect, or the system did not reliably make an edit when applicable (e.g. always hyphenating when appropriate):

\begin{quote}
    \small
    \vspace{-2mm}
    ``Because I wasn't sure I can trust the suggestions (because I saw several incorrect ones) so it took me longer to think/check whether the suggestion is right. And I have to read the entire sentence again anyway to check for other errors the suggestion didn't catch...would only work if I knew 100\% that the suggestions are always right''
    \vspace{-2mm}
\end{quote}

These comments are in concordance with our quantitative findings. Perhaps unlike other assistive applications, it is not enough to only have high precision: if reviewers cannot trust that the system has caught most or all errors, they will not save time as they still have to read the entire sentence carefully. Conversely, a high-recall, low-precision system is not only distracting, but also leads reviewers to be suspicious of whether suggestions are correct in general. In general, future TEC systems must manage this balance of precision and recall for user trust.

\subsubsection{Use Cases for TEC}
Many reviewers highlighted scenarios where trustworthy TEC systems could be particularly useful.

Two reviewers said TEC is helpful for  \textbf{corrections and typos}, similar to the use cases for GEC in the wild \citep{omelianchuk-etal-2021-text}:
\begin{quote}
    \small
    \vspace{-2mm}
     ``If the tool would manage to reliably show missing punctuation marks, or numbers, or that the translation contains different numbers than the source, that would be helpful and save time.''; ``recurring mistakes''
    \vspace{-2mm}
\end{quote}
On the other hand, three reviewers mentioned that they hoped such a system would make more substantial corrections in order to save a non-negligible amount of time, although of course these edits may come at the expense of precision:
\begin{quote}
    \small
    \vspace{-2mm}
    ``There were not many suggestions, and they only offered small improvements...Not clear whether I would save time or not.''
    \vspace{-2mm}
\end{quote}

Three reviewers commented that a TEC system could be a \textbf{memory aid or substitute for researching client-specific requirements}, which is often an intensive part of the production translation process. One reviewer pointed out it could be particularly useful as an \emph{instructive} tool for translators who are new to a client:
\begin{quote}
    \small
    \vspace{-2mm}
    ``if I am new to an account and don't yet know whether this client wants hyphens or not (always an issue with German). So usually I have to research... (or guess), but if the QA suggestions knew this client's preference and would tell me, that would save me time.''
    \vspace{-2mm}
\end{quote}

Finally, three reviewers commented that it could be useful as an \textbf{attention-directing tool} by making them aware of what errors they might look out for, especially in repetitive content where it may be easy to miss details:
\begin{quote}
    \small
    \vspace{-2mm}
    ``makes you more sensitive for spotting similar errors''; ``makes you aware of what kind of errors to look for in upcoming segments''; ``maybe it helps with [repetitive sentences] that you would otherwise just quickly glance at.''
    \vspace{-2mm}
\end{quote}
\section{Conclusion \& Future Work}

We introduced the task of translation error correction (TEC) and released the \datasetname{} corpus to study automatic correction of human translations, consisting of three TEC datasets across varying domains. In our analysis of TEC data, we showed how the errors that humans make differ from those made by MT systems, suggesting that this task warrants different approaches from those previously studied in the task of automatic post-editing. We confirm this empirically by proposing a synthetic data generation procedure that more closely matches the distribution of human translation errors and showing that our TEC model, pre-trained on this data, consistently outperforms models developed for APE, as well as those for MT and GEC. Finally, we showed how our TEC system is helpful to real humans, assisting professional reviewers and leading them to produce higher quality reviewed translations.

Future work may improve on our TEC system by investigating how to leverage the strengths of recent MT systems (e.g. for initializing systems or proposing edits) or developing more sophisticated synthetic data generation techniques (e.g. using the source sentence or linguistic knowledge). Beyond our benchmark, it would be interesting to apply TEC systems to other settings in which human translation errors appear, e.g., to correct translations written by language learners, denoise MT training sets, or clean up MT evaluation sets.

From the perspective of human-AI interaction, TEC presents a real-world use case and testbed to study how to assist experts with modern NLP systems, hinting at the opportunity to combine the best of humans and machines.

\section*{Acknowledgments}
\vspace{-2mm}
We thank Sai Gouravajhala, Yunsu Kim, Eric Wallace, and the other members of the Lilt research team and Berkeley NLP group for helpful discussion and feedback. We thank Morgan Raymond and Spence Green for their support in releasing the dataset. Finally, we are grateful to the professional translators who annotated the dataset and participated in the user study.

\bibliography{anthology,custom}

\begin{thebibliography}{24}
\expandafter\ifx\csname natexlab\endcsname\relax\def\natexlab#1{#1}\fi

\bibitem[{Bryant et~al.(2017)Bryant, Felice, and
  Briscoe}]{bryant-etal-2017-automatic}
Christopher Bryant, Mariano Felice, and Ted Briscoe. 2017.
\newblock \href {https://doi.org/10.18653/v1/P17-1074} {Automatic annotation
  and evaluation of error types for grammatical error correction}.
\newblock In \emph{Proceedings of the 55th Annual Meeting of the Association
  for Computational Linguistics (Volume 1: Long Papers)}, pages 793--805,
  Vancouver, Canada. Association for Computational Linguistics.

\bibitem[{Chollampatt et~al.(2020)Chollampatt, Susanto, Tan, and
  Szymanska}]{chollampatt2020automatic}
Shamil Chollampatt, Raymond~Hendy Susanto, Liling Tan, and Ewa Szymanska. 2020.
\newblock \href {https://doi.org/10.18653/v1/2020.emnlp-main.217} {Can
  automatic post-editing improve {NMT}?}
\newblock In \emph{Proceedings of the 2020 Conference on Empirical Methods in
  Natural Language Processing (EMNLP)}, pages 2736--2746, Online. Association
  for Computational Linguistics.

\bibitem[{Correia and Martins(2019)}]{correira2019apebert}
Gon{\c{c}}alo~M. Correia and Andr{\'e} F.~T. Martins. 2019.
\newblock \href {https://www.aclweb.org/anthology/P19-1292} {A simple and
  effective approach to automatic post-editing with transfer learning}.
\newblock In \emph{Proceedings of the 57th Annual Meeting of the Association
  for Computational Linguistics}, Florence, Italy. Association for
  Computational Linguistics.

\bibitem[{Dahlmeier and Ng(2012)}]{dahlmeier2012m2}
Daniel Dahlmeier and Hwee~Tou Ng. 2012.
\newblock \href {https://www.aclweb.org/anthology/N12-1067} {Better evaluation
  for grammatical error correction}.
\newblock In \emph{Proceedings of the 2012 Conference of the North {A}merican
  Chapter of the Association for Computational Linguistics: Human Language
  Technologies}, pages 568--572, Montr{\'e}al, Canada. Association for
  Computational Linguistics.

\bibitem[{DePalma(2021)}]{depalma-2021-csa}
Donald~A. DePalma. 2021.
\newblock \href
  {https://csa-research.com/Blogs-Events/Blog/ArticleID/785/The-Language-Sector-in-Eight-Charts}
  {The language sector in eight charts}.
\newblock Accessed: 2022-05-01.

\bibitem[{Freitag et~al.(2021)Freitag, Foster, Grangier, Ratnakar, Tan, and
  Macherey}]{freitag2021experts}
Markus Freitag, George Foster, David Grangier, Viresh Ratnakar, Qijun Tan, and
  Wolfgang Macherey. 2021.
\newblock \href {http://arxiv.org/abs/2104.14478} {Experts, errors, and
  context: {A} large-scale study of human evaluation for machine translation}.

\bibitem[{Gupta et~al.(2021)Gupta, Juneja, Nelakanti, and
  Chatterjee}]{Gupta2021DetectingOE}
Prabhakar Gupta, Ridha Juneja, Anil~Kumar Nelakanti, and Tamojit Chatterjee.
  2021.
\newblock Detecting over/under-translation errors for determining adequacy in
  human translations.
\newblock \emph{ArXiv}, abs/2104.00267.

\bibitem[{Hansen(2009)}]{hansen2009errors}
Gyde Hansen. 2009.
\newblock \href
  {https://gydehansen.dk/media/59/a-classification-of-errors-in-translation-and-revisionpeter-lang.pdf}
  {A classification of errors in translation and revision}.
\newblock In \emph{{CIUTI}-Forum: Enhancing Translation Quality: Ways, Means,
  Methods.}

\bibitem[{Junczys-Dowmunt and Grundkiewicz(2016)}]{junczys-dowmunt-2016-logape}
Marcin Junczys-Dowmunt and Roman Grundkiewicz. 2016.
\newblock \href {https://doi.org/10.18653/v1/W16-2378} {Log-linear combinations
  of monolingual and bilingual neural machine translation models for automatic
  post-editing}.
\newblock In \emph{Proceedings of the First Conference on Machine Translation:
  Volume 2, Shared Task Papers}, pages 751--758, Berlin, Germany. Association
  for Computational Linguistics.

\bibitem[{Junczys-Dowmunt et~al.(2018)Junczys-Dowmunt, Grundkiewicz, Guha, and
  Heafield}]{junczys-dowmunt-2018-lowresourcegec}
Marcin Junczys-Dowmunt, Roman Grundkiewicz, Shubha Guha, and Kenneth Heafield.
  2018.
\newblock \href {https://doi.org/10.18653/v1/N18-1055} {Approaching neural
  grammatical error correction as a low-resource machine translation task}.
\newblock In \emph{Proceedings of the 2018 Conference of the North {A}merican
  Chapter of the Association for Computational Linguistics: Human Language
  Technologies, Volume 1 (Long Papers)}, pages 595--606, New Orleans,
  Louisiana. Association for Computational Linguistics.

\bibitem[{Kingma and Ba(2015)}]{kingma-2015-adam}
Diederik~P. Kingma and Jimmy Ba. 2015.
\newblock \href {http://arxiv.org/abs/1412.6980} {Adam: {A} method for
  stochastic optimization}.
\newblock In \emph{3rd International Conference on Learning Representations,
  {ICLR} 2015, San Diego, CA, USA, May 7-9, 2015, Conference Track
  Proceedings}.

\bibitem[{Koehn et~al.(2007)Koehn, Hoang, Birch, Callison-Burch, Federico,
  Bertoldi, Cowan, Shen, Moran, Zens, Dyer, Bojar, Constantin, and
  Herbst}]{koehn-etal-2007-moses}
Philipp Koehn, Hieu Hoang, Alexandra Birch, Chris Callison-Burch, Marcello
  Federico, Nicola Bertoldi, Brooke Cowan, Wade Shen, Christine Moran, Richard
  Zens, Chris Dyer, Ond{\v{r}}ej Bojar, Alexandra Constantin, and Evan Herbst.
  2007.
\newblock \href {https://aclanthology.org/P07-2045} {{M}oses: Open source
  toolkit for statistical machine translation}.
\newblock In \emph{Proceedings of the 45th Annual Meeting of the Association
  for Computational Linguistics Companion Volume Proceedings of the Demo and
  Poster Sessions}, pages 177--180, Prague, Czech Republic. Association for
  Computational Linguistics.

\bibitem[{Lopes et~al.(2019)Lopes, Farajian, Correia, Tr{\'e}nous, and
  Martins}]{lopes-2019-unbabelbert}
Ant{\'o}nio~V. Lopes, M.~Amin Farajian, Gon{\c{c}}alo~M. Correia, Jonay
  Tr{\'e}nous, and Andr{\'e} F.~T. Martins. 2019.
\newblock \href {https://doi.org/10.18653/v1/W19-5413} {Unbabel{'}s submission
  to the {WMT}2019 {APE} shared task: {BERT}-based encoder-decoder for
  automatic post-editing}.
\newblock In \emph{Proceedings of the Fourth Conference on Machine Translation
  (Volume 3: Shared Task Papers, Day 2)}, pages 118--123, Florence, Italy.
  Association for Computational Linguistics.

\bibitem[{Napoles et~al.(2015)Napoles, Sakaguchi, Post, and
  Tetreault}]{napoles2015gleu}
Courtney Napoles, Keisuke Sakaguchi, Matt Post, and Joel Tetreault. 2015.
\newblock \href {https://doi.org/10.3115/v1/P15-2097} {Ground truth for
  grammatical error correction metrics}.
\newblock In \emph{Proceedings of the 53rd Annual Meeting of the Association
  for Computational Linguistics and the 7th International Joint Conference on
  Natural Language Processing (Volume 2: Short Papers)}, pages 588--593,
  Beijing, China. Association for Computational Linguistics.

\bibitem[{Negri et~al.(2018)Negri, Turchi, Chatterjee, and
  Bertoldi}]{negri-etal-2018-escape}
Matteo Negri, Marco Turchi, Rajen Chatterjee, and Nicola Bertoldi. 2018.
\newblock \href {https://aclanthology.org/L18-1004} {{ESCAPE}: a large-scale
  synthetic corpus for automatic post-editing}.
\newblock In \emph{Proceedings of the Eleventh International Conference on
  Language Resources and Evaluation ({LREC} 2018)}, Miyazaki, Japan. European
  Language Resources Association (ELRA).

\bibitem[{Omelianchuk et~al.(2021)Omelianchuk, Raheja, and
  Skurzhanskyi}]{omelianchuk-etal-2021-text}
Kostiantyn Omelianchuk, Vipul Raheja, and Oleksandr Skurzhanskyi. 2021.
\newblock \href {https://aclanthology.org/2021.bea-1.2} {{T}ext
  {S}implification by {T}agging}.
\newblock In \emph{Proceedings of the 16th Workshop on Innovative Use of NLP
  for Building Educational Applications}, pages 11--25, Online. Association for
  Computational Linguistics.

\bibitem[{Sakaguchi et~al.(2016)Sakaguchi, Napoles, Post, and
  Tetreault}]{sakaguchi2016gleufluency}
Keisuke Sakaguchi, Courtney Napoles, Matt Post, and Joel Tetreault. 2016.
\newblock \href {https://doi.org/10.1162/tacl_a_00091} {Reassessing the goals
  of grammatical error correction: Fluency instead of grammaticality}.
\newblock \emph{Transactions of the Association for Computational Linguistics},
  4:169--182.

\bibitem[{Sennrich et~al.(2016)Sennrich, Haddow, and Birch}]{sennrich-2016-bpe}
Rico Sennrich, Barry Haddow, and Alexandra Birch. 2016.
\newblock \href {https://doi.org/10.18653/v1/P16-1162} {Neural machine
  translation of rare words with subword units}.
\newblock In \emph{Proceedings of the 54th Annual Meeting of the Association
  for Computational Linguistics (Volume 1: Long Papers)}, pages 1715--1725,
  Berlin, Germany. Association for Computational Linguistics.

\bibitem[{Simard et~al.(2007)Simard, Ueffing, Isabelle, and
  Kuhn}]{simard-etal-2007-rule}
Michel Simard, Nicola Ueffing, Pierre Isabelle, and Roland Kuhn. 2007.
\newblock \href {https://aclanthology.org/W07-0728} {Rule-based translation
  with statistical phrase-based post-editing}.
\newblock In \emph{Proceedings of the Second Workshop on Statistical Machine
  Translation}, pages 203--206, Prague, Czech Republic. Association for
  Computational Linguistics.

\bibitem[{Specia and Shah(2014)}]{specia-shah-2014-predicting}
Lucia Specia and Kashif Shah. 2014.
\newblock \href {https://aclanthology.org/2014.amta-researchers.22} {Predicting
  human translation quality}.
\newblock In \emph{Proceedings of the 11th Conference of the Association for
  Machine Translation in the Americas: MT Researchers Track}, pages 288--300,
  Vancouver, Canada. Association for Machine Translation in the Americas.

\bibitem[{Vaswani et~al.(2017)Vaswani, Shazeer, Parmar, Uszkoreit, Jones,
  Gomez, Kaiser, and Polosukhin}]{vaswani17transformer}
Ashish Vaswani, Noam Shazeer, Niki Parmar, Jakob Uszkoreit, Llion Jones,
  Aidan~N Gomez, \L~ukasz Kaiser, and Illia Polosukhin. 2017.
\newblock \href
  {https://proceedings.neurips.cc/paper/2017/file/3f5ee243547dee91fbd053c1c4a845aa-Paper.pdf}
  {Attention is all you need}.
\newblock In \emph{Advances in Neural Information Processing Systems},
  volume~30.

\bibitem[{Yuan and Sharoff(2020)}]{yuan-sharoff-2020-sentence}
Yu~Yuan and Serge Sharoff. 2020.
\newblock \href {https://aclanthology.org/2020.lrec-1.229} {Sentence level
  human translation quality estimation with attention-based neural networks}.
\newblock In \emph{Proceedings of the 12th Language Resources and Evaluation
  Conference}, pages 1858--1865, Marseille, France. European Language Resources
  Association.

\bibitem[{Zenkel et~al.(2019)Zenkel, Wuebker, and DeNero}]{zenkel2019adding}
Thomas Zenkel, Joern Wuebker, and John DeNero. 2019.
\newblock \href {https://arxiv.org/abs/1901.11359} {Adding interpretable
  attention to neural translation models improves word alignment}.
\newblock \emph{arXiv preprint arXiv:1901.11359}.

\bibitem[{Zhao et~al.(2019)Zhao, Wang, Shen, Jia, and Liu}]{zhao2019copygec}
Wei Zhao, Liang Wang, Kewei Shen, Ruoyu Jia, and Jingming Liu. 2019.
\newblock \href {https://doi.org/10.18653/v1/N19-1014} {Improving grammatical
  error correction via pre-training a copy-augmented architecture with
  unlabeled data}.
\newblock In \emph{Proceedings of the 2019 Conference of the North {A}merican
  Chapter of the Association for Computational Linguistics: Human Language
  Technologies, Volume 1 (Long and Short Papers)}, pages 156--165, Minneapolis,
  Minnesota. Association for Computational Linguistics.

\end{thebibliography}
\bibliographystyle{acl_natbib}

\appendix
\clearpage
\section{Transformer Architecture Background and Model Description}
\label{sec:appendix-model-description}
\subsection{Transformer Architecture}

The neural models implemented in this work are based on the
self-attentional Transformer architecture \citep{vaswani17transformer}. Formally, given 
a sequence of source tokens (encoded as one-hot vectors) $\V{s}_{1 \dots J} = (s_1, \dots, s_J)$, $s_j \in \M{V}$, the goal is to predict a sequence of target tokens $\V{t'}_{1 \dots I'} = (t'_1, \dots, t'_{I'})$, $t'_i \in \M{V}$, that is a translation of the source sequence, where $\M{V}$ is the vocabulary.
The model has two main components, the \emph{encoder} and the \emph{decoder}. The encoder transforms the source sequence
$\V{s}_{1 \dots J}$ into a sequence of hidden states by first mapping each individual token into a continuous embedding space, adding a
positional embedding and then processing it through a sequence
of self-attention and feed-forward layers:
\begin{align}
\V{x}_{1 \dots J} &= \M{E} \V{s}_{1 \dots J} + \V{p}_{1 \dots J} \\
\V{h}_{1 \dots J}^{\text{enc}} &= \text{encoder}(\V{x}_{1 \dots J}), \label{eq_encoder}
\end{align}
where $x_j \in \M{V}, j \in (1, \dots, J)$, $\M{E}$ is the embedding matrix for vocabulary $\M{E}$ and $\V{p}_{1 \dots J}$ is the sequence of positional embeddings described in Sec. 3.5 of \citep{vaswani17transformer}.
At a given time step $i$, the decoder defines a probability distribution $P_i$ over all vocabulary items in $\M{V}$:
\begin{align}
\V{y'}_{1 \dots i-1} &= \M{E} \V{t'}_{1 \dots i-1} + \V{p}_{1 \dots i-1}\\
\V{h}_i^{\text{dec}} &= \text{decoder}(\V{y'}_{1 \dots i-1}, \V{h}_{1 \dots J}^{\text{enc}}) \label{eq_decoder} \\ 
P_i(t'_i) &= \text{softmax}(\V{h}_i^{\text{dec}} \M{E}^{\top})
\end{align}
where we assume a single shared vocabulary $\M{V}$ and embedding matrix $\M{E}$. 
At training time we optimize the cross-entropy loss 
\begin{align}
    \mathcal{L}_{\text{CE}}(P) = - \sum_{i} \log(P_i(t'_i)).
\end{align}

\subsection{Dual-Source Encoder-Decoder Model}

Given an additional input sequence $\V{t}_{1 \dots I} = (t_1, \dots, t_I)$. the dual-source model used for the APE and \tasknameabr{} models is implemented by independently projecting $\V{t}_{1 \dots I}$ into the embedding space, adding an offset vector $\V{o}$ and concatenating the embedding sequences. Equations \ref{eq_encoder} and \ref{eq_decoder} are rewritten as
\begin{align}
\V{y}_{1 \dots I} &= \M{E} \V{t}_{1 \dots I} + \V{p}_{1 \dots I} + \V{o} \\
\V{h}_{1 \dots (J + I)}^{\text{enc}} &= \text{encoder}([\V{x}_{1 \dots J};\V{y}_{1 \dots I}])\\
\V{h}_i^{\text{dec}} &= \text{decoder}(\V{y'}_{1 \dots i-1}, \V{h}_{1 \dots (J+I)}^{\text{enc}}),
\end{align}
where $\V{o}$ is a single learned vector that is broadcast to all positions $i \in (1, \dots, I)$ and $[\cdot;\cdot]$ denotes the concatenate operation.

\subsection{Copy-Attention Mechanism}

The new output probability distribution for the next target token $P_i(t_i')$ is a weighted sum of the probability of generating and the probability of copying token $t_i'$:
\begin{align}
    \hat{P}_i(t'_i) = (1-\alpha_i^{\text{copy}}) P_i(t'_i) + \alpha_i^{\text{copy}} P_i^{\text{copy}}(t'_i),
\end{align}
where the copy probabilities are calculated from the attention matrix of an additional encoder-decoder attention layer that is added on top of the final decoder layer, $\M{A}_i$:
\begin{align}
    P_i^{\text{copy}}(t'_i) = \text{softmax}(\M{A}_i)
\end{align}
The copy probability weight $\alpha_i^{\text{copy}}$ is determined with the attention context vector $\V{c}_i$, computed as a weighted sum of the attention values (i.e. linearly transformed encoder states) where the weights are defined by $\M{A}_i$:
\begin{align}
    \alpha_i^{\text{copy}} = \text{sigmoid}(\M{W}^{\top} \V{c}_i).
\end{align}
This copy-attention layer applies a source-side mask so that it only attends to the positions $(J+1, \dots, J+I)$ that correspond to the second input sequence $\V{t}_{1 \dots I}$, and its implementation follows \citet{zenkel2019adding}. 
In particular, it uses a single attention head, no skip connection, and contains a separate output layer that predicts the target word based on its context vector with probability distribution $P_i^{\text{align}}(\cdot)$. 
At training time both output layers are optimized jointly by defining the overall loss $\mathcal{L}$ as the weighted sum of both
cross-entropy losses:
\begin{align}
    \mathcal{L} = \mathcal{L}_{\text{CE}}(\hat{P}) + \lambda \mathcal{L}_{\text{CE}}(P^{\text{align}})
\end{align}
$\lambda$ is set to $0.05$ in all experiments. We further apply source-word dropout \citep{junczys-dowmunt-2018-lowresourcegec}, setting the full embedding vector for words in \pert{} to $1/p_{\text{src}}$ with probability $p_{\text{src}} = 0.05$.

\section{Results without finetuning}
\label{sec:appendix-base-results}
See \tabref{tab:appendix-no-finetuning}.
\begin{table*}[!ht]
\centering
\small
\begin{tabular}{lccccccccc}\toprule
& \multicolumn{3}{c}{\asics{}} & \multicolumn{3}{c}{\emerson{}} & \multicolumn{3}{c}{\digitalocean{}} \\ 
\cmidrule(lr){2-4}\cmidrule(lr){5-7}\cmidrule(lr){8-10}
\bf Model & Prec. & Rec. & F$_{0.5}$ & Prec. & Rec. & F$_{0.5}$ & Prec. & Rec. & F$_{0.5}$ \\\midrule
MT      & 0.8	& \bf 12.2 &	1.0	& 1.6	&  \bf 16.9	& 2.0 & 0.5 &	\bf 23.4 &	0.6  \\
GEC     & 5.9	& 1.4	& 3.6	& 0.7	& 0.6	& 0.7 & 0.3	& 1.2	& 0.3      \\
APE    & 2.7 & 	7.3 & 	3.1 &	3.6 &	3.3 &	3.5  & 0.4	& 5.0	& 0.5  \\
BERT-APE & 0.5 &	5.9 &	0.6 &	0.1 &	2.2 &	0.1 & 0.3 &	9.2 & 0.4\\
\midrule
\modelname{} & \bf 41.7 & 1.7 & \bf 7.5	& \bf 12.2	& 4.2	& \bf 8.8 & \bf 5.3 & 2.5	& \bf 4.3 \\
\bottomrule
\end{tabular}
\caption{Results without finetuning.}
\label{tab:appendix-no-finetuning}
\end{table*}

\section{Examples of System Output}
\label{sec:appendix-system-outputs}
See \tabref{tab:appendix-sys-outputs1}, \tabref{tab:appendix-sys-outputs2}, \tabref{tab:appendix-sys-outputs3}.

\begin{table*}[t]
\begin{tabular}{lA}\toprule
Type: & \multicolumn{1}{l}{Monolingual: technical} \\
\src{}: & Do your feet roll inwards when running? \\
\pert{}: & K\remove{I}ppen deine Füße beim Laufen nach innen ? \\
Reference \tgt{}: & K\add{i}ppen deine Füße beim Laufen nach innen ?\\\midrule
\textbf{Model} & \multicolumn{1}{l}{\textbf{Predicted \tgt{}}} \\\hline
\bf MT & \remove{KIppen}\add{Rollen} deine Füße beim Laufen nach innen ?  \\
\bf APE & Correctly predicts \tgt{} \\
\bf BERT-APE & No change to \pert{} \\
\bf GEC & Correctly predicts \tgt{} \\
\bf TEC (ours) & Correctly predicts \tgt{} \\\bottomrule
\end{tabular}
\caption{A monolingual technical error the APE, GEC and TEC models edit correctly.}
\label{tab:appendix-sys-outputs1}
\end{table*}

\begin{table*}
\begin{tabular}{lA}\toprule
Type: & \multicolumn{1}{l}{Monolingual: technical} \\
\src{}: & Run further than you thought possible in the GEL-NIMBUS ™ 21 LS running shoe for men by ASICS - packed full of plush cushioning and special technologies to respond to your natural stride. \\
\pert{}: & Mit \remove{mit} dem ASICS GEL-NIMBUS ™ 21 LS Laufschuh für Herren mit weicher Dämpfung und speziellen Technologien , durch die sich der Schuh an deine natürlichen Schritte anpasst , läufst du weiter , als du es selbst je für möglich gehalten hast .\\
Reference \tgt{}: & Mit dem ASICS GEL-NIMBUS ™ 21 LS Laufschuh für Herren mit weicher Dämpfung und speziellen Technologien , durch die sich der Schuh an deine natürlichen Schritte anpasst , läufst du weiter , als du es selbst je für möglich gehalten hast .\\\midrule
\textbf{Model} & \multicolumn{1}{l}{\textbf{Predicted \tgt{}}} \\\hline
\bf MT & \remove{Mit mit dem}\add{Laufe weiter , als du es im} GEL-NIMBUS ™ 21 LS Laufschuh für Herren  \add{ von ASICS für möglich gehalten hättest -} mit weicher Dämpfung und speziellen Technologien , \remove{durch die sich der Schuh an deine natürlichen Schritte anpasst , läufst du weiter , als du es} \remove{ selbst je für möglich gehalten hast .}\add{um auf deinen natürlichen Schritt zu reagieren .}  \\
\bf APE & No change to \pert{} \\
\bf BERT-APE & Correctly predicts \tgt{} \\
\bf GEC & Correctly predicts \tgt{} \\
\bf TEC (ours) & Correctly predicts \tgt{} \\\bottomrule

\end{tabular}
\caption{A monolingual technical error the BERT-APE, GEC and TEC models edit correctly.}
\label{tab:appendix-sys-outputs2}
\end{table*}

\begin{table*}
\begin{tabular}{lA}\toprule
Type: & \multicolumn{1}{l}{Bilingual} \\
\src{}: & The DUOMAX ™ midsole offers smooth overpronation control by combining two different density materials to reduce the risk of flat feet and bunions.\\
\pert{}: & Die DUOMAX ™ -Mittelsohle bietet mühelos Halt bei Überpronation , indem zwei unterschiedliche Dichtematerialien kombiniert werden , um das Risiko von \remove{flachen Füßen} und \remove{Fußballen} zu verringern . \\
Reference \tgt{}: & Die DUOMAX ™ -Mittelsohle bietet mühelos Halt bei Überpronation , indem zwei unterschiedliche Dichtematerialien kombiniert werden , um das Risiko von \add{Plattfüßen} und \add{Ballenzehen} zu verringern . \\ \midrule
\textbf{Model} & \multicolumn{1}{l}{} \\\hline
\bf MT & Die DUOMAX ™ -Mittelsohle bietet \remove{mühelos Halt bei}\add{eine reibungslose} Überpronation , indem \add{sie} zwei unterschiedliche Dichtematerialien kombiniert , um das Risiko von flachen Füßen und \remove{Fußballen}\add{Bündchen} zu verringern . \\
\bf APE &  No change to \pert{} \\
\bf BERT-APE & Die DUOMAX ™ -Mittelsohle bietet mühelos Halt bei Überpronation , indem zwei unterschiedliche Dichtematerialien kombiniert werden , um das Risiko von flachen Füßen und \remove{Fußballen}\add{Baseballen} zu verringern . \\
\bf GEC & No change to \pert{} \\
\bf TEC (ours) & No change to \pert{} \\\bottomrule

\end{tabular}
\caption{A bilingual error all models fail to edit correctly.}
\label{tab:appendix-sys-outputs3}
\end{table*}

\section{Full User Study Results}
\label{sec:appendix-all-user-study}
\begin{table*}[t]
\centering
\small
\begin{tabular}{@{}lllll@{}}
\toprule
                                       & \begin{tabular}[c]{@{}l@{}}Suggestion\\ Hidden\end{tabular} & \begin{tabular}[c]{@{}l@{}}Suggestion\\ Shown\end{tabular} & \begin{tabular}[c]{@{}l@{}}Suggestion Shown\\ and Accepted\end{tabular} & \begin{tabular}[c]{@{}l@{}}Suggestion Shown\\ and Declined\end{tabular} \\ \midrule
Review Time (median)                   & 34.0475 sec                                                 & 33.103 sec                                                 & 26.606 sec                                                              & 50.524 sec                                                              \\
Review Time (length-norm, median)      & 361 ms/char                                                 & 367 ms/char                                                & 328 ms/char                                                             & 841 ms/char                                                             \\
Inserts (median)                       & 1.5 chars                                                   & 0 chars                                                    & 0 chars                                                                 & 4 chars                                                                 \\
Inserts (length-norm, median)          & 0.0248                                                      & 0                                                          & 0                                                                       & 0.0345                                                                  \\
Deletes (median)                       & 2 chars                                                     & 0 chars                                                    & 0 chars                                                                 & 4 chars                                                                 \\
Deletes (length-norm, median)          & 0.0294                                                      & 0                                                          & 0                                                                       & 0.0375                                                                  \\
Inserts+Deletes (median)               & 4 chars                                                     & 0 chars                                                    & 0 chars                                                                 & 9 chars                                                                 \\
Inserts+Deletes (length-norm, median)  & 0.0625                                                      & 0                                                          & 0                                                                       & 0.0625                                                                  \\
Levenshtein Dist (median)              & 2 chars                                                     & 1 char                                                     & 1 char                                                                  & 6 chars                                                                 \\
Levenshtein Dist (length-norm, median) & 0.0347                                                      & 0.0185                                                     & 0.0176                                                                  & 0.0426                                                                  \\ \bottomrule
\end{tabular}
\caption{All data from our user study about review times, number of characters the user inserted and deleted, and final Levenshtein distances from the original. Data shown are medians (raw and length-normalized) across the segments, based on whether the suggestion was hidden or shown. ``Suggestion Shown'' is further broken down according to whether the user accepted or declined the suggestion.}
\label{tab:appendix-user-study-results}
\end{table*}
See \tabref{tab:appendix-user-study-results}.

\end{document}